\documentclass{article}

\usepackage[utf8]{inputenc}
\usepackage{amsmath}
\usepackage{amssymb}
\usepackage{booktabs}
\usepackage{graphicx}
\usepackage{svg}
\usepackage[a4paper,margin=1in]{geometry}
\usepackage{microtype}
\usepackage[hidelinks]{hyperref}
\usepackage{caption}
\usepackage{subcaption}
\usepackage[numbers]{natbib}
\title{
BCMT: Blockwise Causal Memory Transformer
}

\author{
Anonymous Authors
}

\author{
Rachid Arezki\\
Independent Researcher\\
France, Nîmes\\
\texttt{rachid.arezki.pro@gmail.com}
}

\date{}

\begin{document}

\maketitle

\begin{abstract}

Transformer architectures rely on dense self-attention to model long-range dependencies, but this mechanism exhibits quadratic complexity with respect to sequence length.

We introduce BCMT (\textit{Blockwise Causal Memory Transformer}), an architecture for long-context language modeling that decouples local token interactions from global context propagation. Dense causal self-attention is applied independently within local blocks, while each block produces an adaptive summary aggregated through an exponential causal memory. This memory is subsequently injected back into the token representations, enabling efficient propagation of long-range contextual information without relying on explicit global attention.

Unlike standard Transformers and recurrent memory architectures, BCMT maintains neither dense interactions between distant tokens nor learned memory states. Its memory mechanism is fully parallelizable and remains compatible with standard implementations of dense self-attention.

Experiments on language modeling with context lengths of up to 1024 tokens show that BCMT achieves validation performance comparable to that of Dense Transformers while significantly improving training throughput and reducing memory consumption. An ablation study further confirms that these improvements arise from the proposed memory mechanism.

These results demonstrate that an exponential causal memory constructed from block summaries provides an effective alternative to dense global attention mechanisms for long-context language modeling.

\end{abstract}

\section{Introduction}

Transformer architectures have become the dominant paradigm for autoregressive language modeling owing to the expressive power of self-attention \cite{vaswani2017attention}. By allowing each token to interact with all preceding tokens, they effectively model complex dependencies over long sequences.

This flexibility, however, comes at a high computational cost. Causal self-attention exhibits quadratic complexity with respect to sequence length, which limits its applicability as the context becomes longer.

Numerous approaches have therefore been proposed as alternatives to dense global attention, including sparse attention mechanisms \cite{zaheer2020bigbird,beltagy2020longformer}, recurrent memory architectures \cite{dai2019transformerxl}, and state-space models \cite{gu2024mamba,sun2023retnet,peng2023rwkv}. Although these approaches improve scalability, they generally rely either on modifying the attention structure or on introducing more complex recurrent mechanisms.

In this work, we propose BCMT (\textit{Blockwise Causal Memory Transformer})\footnote{Official implementation: \url{https://github.com/rachidlabs/BCMT}}, an architecture based on a different principle. The central idea is to decouple the modeling of local dependencies from global context propagation. Token interactions are captured by dense causal self-attention applied independently within local blocks, while long-range dependencies are propagated through an exponential causal memory constructed from adaptive block summaries.

The resulting memory representations are then injected back into the token representations through a gating mechanism, enabling the propagation of global contextual information without relying on explicit attention between tokens belonging to different blocks. Unlike recurrent memory architectures, the memory mechanism of BCMT relies neither on learned hidden states nor on persistent memory tokens. Instead, it is obtained through a fully parallelizable exponential causal aggregation that remains compatible with standard GPU implementations of dense self-attention.

We evaluate BCMT on language modeling tasks with context lengths of up to 1024 tokens. The results show that the proposed architecture achieves validation performance comparable to that of a Dense Transformer while significantly improving training throughput and reducing memory consumption. An ablation study further confirms that these improvements arise from the proposed memory mechanism rather than from the sequence partitioning alone.

Our main contributions are as follows:

\begin{itemize}

\item We propose BCMT, an architecture that explicitly decouples local token interactions from global context propagation through an inter-block exponential causal memory.

\item We introduce an exponential causal memory mechanism constructed from adaptive block summaries, enabling efficient modeling of long-range dependencies without explicit global attention.

\item We experimentally demonstrate that BCMT preserves modeling performance comparable to that of a Dense Transformer while significantly improving computational efficiency, training throughput, and memory consumption.

\end{itemize}

\section{Related Work}

\subsection{Scaling Self-Attention to Long Sequences}

The success of Transformer architectures largely stems from the expressive power of dense self-attention \cite{vaswani2017attention}. However, the quadratic complexity of this mechanism with respect to sequence length quickly limits its applicability to long-context settings.

A first family of approaches reduces this cost through sparse attention patterns. Longformer restricts attention to sliding local windows complemented by a limited number of global tokens \cite{beltagy2020longformer}, while BigBird combines local, random, and global connections to approximate a dense attention graph \cite{zaheer2020bigbird}. These methods reduce the computational cost while preserving explicit interactions between distant tokens.

A second family relies on approximations of the attention operator itself. Linformer projects keys and values into a lower-dimensional space \cite{wang2020linformer}, whereas Performer replaces softmax attention with an approximation based on random feature mappings \cite{choromanski2021performer}. These approaches improve scalability at the cost of modifying the original formulation of self-attention.

Finally, more recent architectures designed for very long contexts exploit structured attention mechanisms. LongNet introduces dilated attention, enabling a substantial increase in the effective context while limiting the number of computed interactions \cite{ding2023longnet}. Despite their differences, all of these approaches retain an attention mechanism that explicitly connects distant tokens.

BCMT follows a different strategy. Attention is strictly confined to independent local blocks, while long-range dependencies are propagated through an exponential causal memory constructed from adaptive block summaries. Consequently, unlike the aforementioned approaches, BCMT neither extends the receptive field nor modifies the structure of the attention graph. Instead, long-range interactions emerge from the progressive propagation of a compact contextual representation across blocks, rather than from explicit attention between distant tokens.

\subsection{Local and Blockwise Sequence Processing}

Another line of research consists in modifying the processing unit of the sequence by replacing global attention with the independent processing of local subsequences. This strategy relies on the assumption that the most informative interactions are predominantly local, thereby reducing the computational cost of self-attention while preserving compatibility with standard dense attention implementations.

BlockBERT is a representative example of this approach. The input sequence is partitioned into contiguous blocks, each processed independently using dense intra-block self-attention \cite{qiu2019blockbert}. This organization reduces the computational cost while naturally leveraging existing GPU implementations.

The main limitation of these architectures is the absence of an explicit communication mechanism between blocks. Long-range dependencies can only be captured by increasing the block size or by introducing additional mechanisms, which progressively reduces the computational benefits of local processing.

BCMT builds upon the principle of local blockwise processing by introducing an explicit context propagation mechanism. Each block is summarized into an adaptive vector, and these summaries are subsequently aggregated through an exponential causal memory to construct a compact representation of the sequence history. Long-range dependencies are therefore propagated across blocks without reintroducing explicit attention between tokens belonging to different blocks.

\subsection{Sequence Compression and Pooling-Based Architectures}

Another family of approaches improves the scalability of Transformer architectures by explicitly reducing the effective sequence length through compression or pooling mechanisms.

Set Transformer introduces learned aggregation operators that efficiently summarize a set of representations into a lower-dimensional latent space \cite{lee2019settransformer}. Funnel Transformer, in contrast, applies hierarchical pooling to progressively reduce the temporal resolution throughout the network, thereby decreasing the computational cost of deeper layers \cite{dai2020funnel}.

More recently, several approaches have investigated dynamically reducing the number of tokens. Token Pooling selects or aggregates intermediate representations to reduce the computational cost of subsequent layers \cite{marin2022tokenpooling}, whereas PoolingFormer partially replaces attention operations with parameterized pooling mechanisms capable of capturing contextual information at low computational cost \cite{yu2022metaformer}.

These approaches share a common principle: they use compressed representations to reduce the complexity of subsequent processing. BCMT also relies on block summaries obtained through adaptive pooling, but with a different objective. The compressed representations neither replace the original tokens nor reduce the sequence length processed by subsequent layers. Instead, they constitute an auxiliary memory channel dedicated to propagating context across blocks.

Consequently, compression is not used as a mechanism for reducing computational cost, but rather as a memory support. The adaptive summaries are aggregated through an exponential causal memory to propagate a compact representation of the sequence history, while the local token representations are preserved and continue to be processed by dense self-attention within each block.

\subsection{Hierarchical and Multi-Scale Sequence Modeling}

Hierarchical representations have long been investigated as a means of modeling dependencies across multiple temporal scales. Early work on recurrent neural networks already explored mechanisms for temporal abstraction and memory compression \cite{elman1990finding,schmidhuber1992learning}. These ideas subsequently evolved into hierarchical recurrent architectures and multi-scale latent sequence models.

More recently, these principles have been adapted to Transformer architectures. Hierarchical Autoregressive Transformers exploit representations operating at different linguistic resolutions \cite{neitemeier2025hierarchical}, while Hierarchical Resolution Transformers introduce multi-scale processing inspired by wavelet decompositions through multiple levels of abstraction \cite{sar2025hrt}. Other works propose hierarchical memory organizations to structure long-term reasoning or the pretraining of large language models \cite{sun2025hmem,pouransari2025hierarchical}.

BCMT shares with these approaches the idea that compact representations can facilitate the modeling of long-range dependencies. However, its objective is fundamentally different. Unlike hierarchical architectures, BCMT neither constructs multiple levels of representation nor builds a hierarchy of progressively coarser resolutions. Instead, it relies on a single level of adaptive block summaries, aggregated through an exponential causal memory to progressively propagate context along the sequence.

Thus, BCMT does not seek to learn a hierarchy of representations, but rather to decouple local token processing from global context propagation. The exponential causal memory acts as a compact context propagation channel between blocks while preserving dense local self-attention within each block.

\subsection{Memory-Augmented and Recurrent Transformer Architectures}

Another major direction for modeling long contexts consists of augmenting Transformer architectures with explicit memory or recurrent mechanisms.

Transformer-XL introduces segment-level recurrence by reusing the hidden states of previous segments, thereby extending the effective context window \cite{dai2019transformerxl}. Compressive Transformer further develops this idea by maintaining compressed historical representations across multiple temporal scales \cite{rae2020compressive}.

More recently, several works have proposed integrating persistent memory directly into Transformer architectures. Recurrent Memory Transformer (RMT) introduces dedicated memory tokens that are propagated across successive segments \cite{bulatov2022recurrent}. Hierarchical Memory Transformer (HMT) organizes these memory representations into multiple hierarchical levels to simultaneously capture short-, medium-, and long-range dependencies \cite{he2024hierarchical}. Along the same line, \textit{Diagonal Batching Unlocks Parallelism in Recurrent Memory Transformers} introduces a computation reordering strategy that enables the parallel execution of RMT-based architectures while exactly preserving their recurrent dynamics. This approach significantly improves the hardware efficiency of recurrent memory architectures without modifying their underlying formulation \cite{sivtsov2025diagonal}.

In parallel, other works have proposed replacing self-attention with learned recurrent dynamics. RetNet \cite{sun2023retnet}, RWKV \cite{peng2023rwkv}, and Mamba \cite{gu2024mamba} model long-range dependencies through recurrently updated internal states rather than explicit interactions between tokens.

BCMT shares with memory-augmented architectures the idea of propagating a compact representation of the context across segments instead of maintaining dense attention over the entire sequence. However, its memory mechanism differs fundamentally from previous approaches. Unlike Transformer-XL, RMT, or HMT, BCMT relies neither on recurrent hidden states nor on persistent memory tokens, and it does not learn any recurrent dynamics.

The memory in BCMT is constructed directly from adaptive block summaries produced after local self-attention. These summaries are then aggregated through a normalized exponential causal memory, yielding a compact and stable representation of the sequence history. This construction is entirely deterministic, fully parallelizable, and independent of any learned recurrent memory state.

Thus, BCMT differs from existing memory-based architectures by replacing recurrent mechanisms with context propagation based on an exponential causal aggregation of block summaries, while preserving dense local self-attention within each block.

\subsection{Positioning of BCMT}
The approaches reviewed above illustrate several strategies for extending the capabilities of Transformer architectures to long-context modeling, including reducing attention connectivity, approximating the attention operator, local blockwise processing, hierarchical representation compression, and the introduction of explicit memory or recurrent mechanisms.

BCMT differs from these approaches by combining two simple principles: dense causal self-attention strictly confined to independent local blocks and an exponential causal memory constructed from adaptive summaries of these blocks. Consequently, long-range dependencies are neither modeled through global attention nor stored in learned recurrent states or persistent memory tokens.

Rather than extending the receptive field of self-attention or maintaining a memory state that evolves throughout the sequence, BCMT progressively propagates a compact contextual representation across blocks. This explicit separation between local token interactions and global context propagation constitutes the central principle of the proposed architecture.

Through this design, BCMT positions itself as an alternative to both global-attention Transformers and recurrent memory architectures by introducing a context propagation mechanism that is fully parallelizable and compatible with standard implementations of dense self-attention.

\section{Method}

BCMT is based on two complementary mechanisms. Dense causal self-attention is applied independently within local blocks to model interactions between nearby tokens. In parallel, an exponential causal memory constructed from adaptive block summaries propagates global context across blocks. This separation preserves strictly local attention while efficiently propagating long-range dependencies.

\subsection{Problem Formulation}

We consider an autoregressive language modeling task over a sequence of length \(T\), represented as

\[
X = (x_1, x_2, \ldots, x_T),
\qquad
x_t \in \mathbb{R}^{D},
\]

where \(D\) denotes the dimensionality of the token representations.

The objective is to model the conditional distribution \(p(x_t \mid x_{<t})\) while preserving autoregressive causality. To reduce the computational cost of self-attention for long sequences, BCMT replaces global attention across all tokens with a combination of local attention and inter-block memory.

\begin{figure}[ht]
\centering
\includesvg[width=\textwidth]{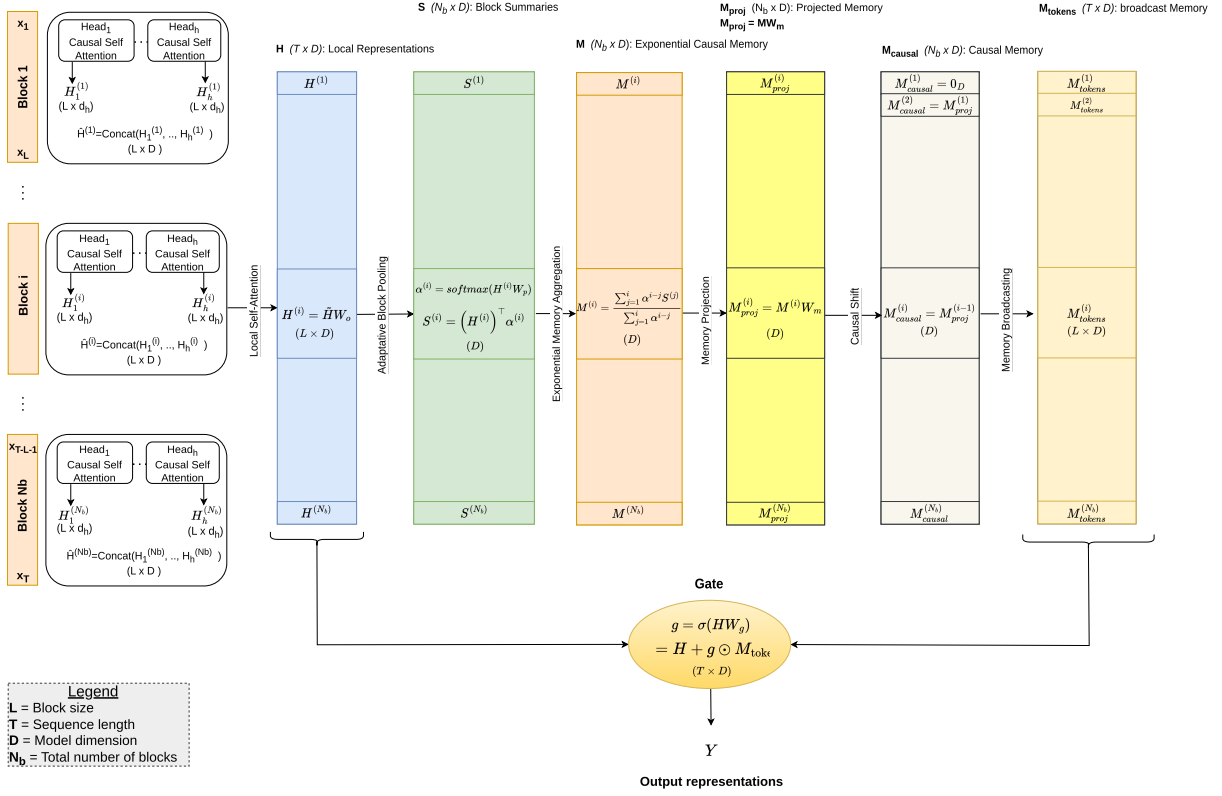}
\caption{Overview of the BCMT architecture.}
\label{fig:bcmt_architecture}
\end{figure}

\subsection{Block Decomposition}

The input sequence is partitioned into \(N_b\) contiguous, non-overlapping blocks of maximum size \(L\):

\[
X
=
\left\{
X^{(1)},
X^{(2)},
\dots,
X^{(N_b)}
\right\},
\]

with

\[
N_b
=
\left\lceil
\frac{T}{L}
\right\rceil.
\]

Each block \(X^{(i)} \in \mathbb{R}^{L_i \times D}\) contains \(L_i \leq L\) tokens. All blocks have size \(L\), except possibly the last one when \(T\) is not a multiple of \(L\).

This decomposition constitutes the fundamental principle of BCMT. Token interactions are restricted to each individual block through dense causal self-attention, while dependencies across blocks are handled by the memory mechanism described in the following sections.

\subsection{Local Multi-Head Causal Attention (\(H\))}

Each block is processed independently using a standard multi-head causal self-attention layer.

For a block

\[
X^{(b)}
\in
\mathbb{R}^{L \times D},
\]

and for each attention head

\[
i \in \{1,\ldots,h\},
\]

where \(D = h\,d_h\), the query, key, and value projections are defined as

\[
Q_i^{(b)}
=
X^{(b)}W_{Q,i},
\qquad
K_i^{(b)}
=
X^{(b)}W_{K,i},
\qquad
V_i^{(b)}
=
X^{(b)}W_{V,i}.
\]

The output of head \(i\) is then given by

\[
H_i^{(b)}
=
\mathrm{softmax}
\left(
\frac{
Q_i^{(b)}
(K_i^{(b)})^{T}
}{
\sqrt{d_h}
}
+
\mathcal{M}
\right)
V_i^{(b)},
\]

where \(\mathcal{M}\) denotes the standard causal mask.

The outputs of all heads are concatenated and projected back into the representation space:

\[
\hat H^{(b)}
=
\mathrm{Concat}
\left(
H_1^{(b)},
\dots,
H_h^{(b)}
\right),
\]

\[
H^{(b)}
=
\hat H^{(b)}W_O,
\qquad
W_O
\in
\mathbb{R}^{D\times D},
\]

with

\[
H^{(b)}
\in
\mathbb{R}^{L\times D}.
\]

The local representations of all blocks are then concatenated along the sequence dimension:

\[
H
=
\begin{bmatrix}
H^{(1)}\\
H^{(2)}\\
\vdots\\
H^{(N_b)}
\end{bmatrix}
\in
\mathbb{R}^{T\times D}.
\]

The tensor \(H\) constitutes the local representation produced by BCMT. It serves as the starting point for constructing the adaptive block summaries, from which the inter-block memory described in the following section is computed.

\subsection{Adaptive Block Summaries (\(S\))}

Starting from the local representations \(H^{(i)}\), each block is summarized by a single contextual vector. Rather than using a uniform average of the token representations, BCMT learns an adaptive weighting that aggregates the content of each block.

For a block

\[
H^{(i)}
\in
\mathbb{R}^{L\times D},
\]

the token scores are obtained through a linear projection:

\[
s^{(i)}
=
H^{(i)}W_p,
\]

where

\[
W_p
\in
\mathbb{R}^{D},
\]

and

\[
s^{(i)}
\in
\mathbb{R}^{L}.
\]

The scores are then normalized using the softmax function:

\[
\alpha^{(i)}
=
\mathrm{softmax}
\left(
s^{(i)}
\right),
\]

with

\[
\alpha^{(i)}
\in
\mathbb{R}^{L}.
\]

The adaptive block summary is then computed as

\[
S^{(i)}
=
\left(H^{(i)}\right)^{\top}
\alpha^{(i)},
\]

where

\[
S^{(i)}
\in
\mathbb{R}^{D}.
\]

The vectors \(S^{(i)}\) provide a compact representation of each block. They are subsequently used as the input to the exponential causal memory mechanism described in the following section.

\subsection{Exponential Causal Memory}

The block summaries $\{S^{(i)}\}_{i=1}^{N_b}$ are used to construct an inter-block memory that propagates context throughout the sequence.

For each block \(i\), we construct a memory vector obtained through a normalized causal exponential aggregation of the summaries observed up to the current block:

\[
M^{(i)}
=
\frac{
\sum_{j=1}^{i}
\alpha^{\,i-j}S^{(j)}
}{
\sum_{j=1}^{i}
\alpha^{\,i-j}
},
\]

where $\alpha\in[0,1)$ denotes the exponential decay factor.

Unlike a uniform average, this aggregation naturally assigns greater weight to more recent summaries while preserving a progressively decreasing contribution from the entire history.

The decay factor is fixed to $\alpha = 0.8$. Preliminary experiments showed that values between 0.7 and 0.9 yield very similar performance. We therefore use \(\alpha=0.8\) throughout all experiments.

The denominator normalizes the weighting coefficients so that they form a convex combination of the block summaries, regardless of the number of blocks.

Each memory vector satisfies

\[
M^{(i)}
\in
\mathbb{R}^{D}.
\]

The memory representations are then projected into the local representation space:

\[
M_{\mathrm{proj}}^{(i)}
=
M^{(i)}W_m,
\]

where

\[
W_m
\in
\mathbb{R}^{D\times D}.
\]

To preserve strict autoregressive causality, the projected memory is shifted by one block:

\[
M_{\mathrm{causal}}^{(i)}
=
M_{\mathrm{proj}}^{(i-1)},
\]

with

\[
M_{\mathrm{causal}}^{(1)}
=
\mathbf{0_D}.
\]

Consequently, block \(i\) receives information exclusively from preceding blocks and can never access its own summary.

The resulting memory tensor is therefore

\[
M_{\mathrm{causal}}
\in
\mathbb{R}^{N_b\times D}.
\]

\subsection{Autoregressive Causality}

BCMT satisfies strict autoregressive causality. No token representation depends on information originating from future positions. This property results from the combination of local self-attention and the inter-block memory mechanism.

\paragraph{Local Self-Attention Causality}

Within each block, representations are computed using standard causal self-attention. The lower triangular mask ensures that each token can attend only to itself and to preceding tokens within the same block.

Consequently, no information from future positions can be accessed through the local attention pathway.

\paragraph{Inter-Block Memory Causality}

The adaptive summary \(S^{(i)}\) is constructed solely from the local representations of block \(i\). The exponential causal memory is then computed from the summaries of all blocks observed up to the current block.

Before being injected into the local representations, this memory is shifted by one block:

\[
M_{\mathrm{causal}}^{(i)}
=
M_{\mathrm{proj}}^{(i-1)},
\]

with

\[
M_{\mathrm{causal}}^{(1)}
=
\mathbf{0_D}.
\]

Consequently, block \(i\) receives contextual information exclusively from blocks

\[
\{1,\ldots,i-1\},
\]

and can never access either its own summary or the summaries of future blocks.

\paragraph{Global Causality}

The autoregressive causality of BCMT therefore relies on three complementary properties:

\begin{itemize}
\item strictly causal self-attention within each block;
\item a memory constructed solely from the summaries of the observed blocks;
\item a causal shift of the memory before its injection into the local representations.
\end{itemize}

Consequently, there exists no computational path through which information from a future token can influence the representation of the current token. BCMT therefore satisfies the causality constraints required for autoregressive language modeling.

\subsection{Memory Broadcast to Tokens}

The memory representation of each block is broadcast to all tokens belonging to that block.

For each block \(i\), the memory vector

\[
M_{\mathrm{causal}}^{(i)}
\in
\mathbb{R}^{D}
\]

is replicated across the \(L\) positions of the block to form a matrix

\[
\widetilde{M}^{(i)}
\in
\mathbb{R}^{L\times D},
\]

defined by

\[
\widetilde{M}^{(i)}_{t,:}
=
M_{\mathrm{causal}}^{(i)},
\qquad
t=1,\ldots,L.
\]

The matrices \(\widetilde{M}^{(i)}\) are then concatenated along the sequence dimension, producing the tensor

\[
M_{\mathrm{tokens}}
\in
\mathbb{R}^{T\times D}.
\]

As a result, every token receives the same contextual representation as the other tokens within its block. This representation is constructed exclusively from preceding blocks, in accordance with the autoregressive causality established in the previous section.

The tensor \(M_{\mathrm{tokens}}\) is then combined with the local representations through the gated fusion mechanism described in the following section.

\subsection{Gated Memory Injection}

The exponential causal memory provides contextual information that complements the local representations produced by self-attention. Its contribution is dynamically modulated through a learned gating mechanism.

A gate vector is computed from the local representations:

\[
G
=
\sigma(HW_g),
\]

where

\[
W_g
\in
\mathbb{R}^{D\times D},
\]

and

\[
G
\in
\mathbb{R}^{T\times D}.
\]

The final representation is then computed as

\[
Y
=
H
+
G
\odot
M_{\mathrm{tokens}},
\]

where \(\odot\) denotes the element-wise product.

The coefficients of \(G\) are computed independently for each token and each representation dimension. They control the amount of contextual information injected into the local representations: when the coefficients are close to zero, the local representation is largely preserved; when they are close to one, the contribution of the exponential causal memory becomes more significant.

This mechanism enables the model to adaptively regulate the use of memory according to the context while preserving local information whenever it is sufficient.

\subsection{Conceptual Interpretation}

BCMT explicitly separates the modeling of local dependencies from that of long-range dependencies.

Interactions between neighboring tokens are modeled directly by local causal self-attention operating independently within each block. More distant dependencies, in contrast, are no longer modeled through explicit interactions between all tokens in the sequence, but through the propagation of a compact memory constructed from adaptive block summaries.

This separation introduces an explicit inductive bias into the architecture. Self-attention is specialized for modeling fine-grained, high-resolution interactions, whereas the exponential causal memory captures a more global and progressively aggregated representation of the sequence history.

From a signal processing perspective, BCMT can be interpreted as a decomposition into two complementary scales. The local pathway preserves the high-frequency components associated with interactions between neighboring tokens, whereas the exponential causal memory conveys lower-frequency information describing the global evolution of the context.

This organization differs from that of global-attention Transformers, in which all dependencies are modeled by a single attention mechanism. In contrast, BCMT distributes these two functions across two complementary mechanisms, each operating at a different temporal resolution.

\subsection{Computational Complexity}
\label{subsec:compcompu}

Let \(T\) denote the sequence length, \(L\) the block size, and
\(D\) the representation dimension. The sequence is partitioned into

\[
N_b=\left\lceil\frac{T}{L}\right\rceil
\]

blocks.

The computational cost of BCMT consists of four successive operations:
local intra-block attention, adaptive block summarization, exponential
causal memory construction, and memory injection into the token
representations.

\paragraph{Local Attention}

Each block is processed independently using dense causal self-attention
with complexity

\[
\mathcal{O}(L^2).
\]

Since the sequence is partitioned into \(N_b\) blocks, the overall
attention complexity becomes

\[
N_b\,\mathcal{O}(L^2)
=
\mathcal{O}(TL).
\]

For a fixed block size, the attention cost therefore grows linearly with
the sequence length.

The memory required to store the attention matrices follows the same
analysis and is also

\[
\mathcal{O}(TL),
\]

compared with

\[
\mathcal{O}(T^2)
\]

for a Dense Transformer with global attention.

\paragraph{Adaptive Block Summaries}

Computing the pooling scores and the weighted summary for each block
requires a single pass over the local representations. Its complexity is

\[
\mathcal{O}(N_bLD)
=
\mathcal{O}(TD),
\]

which remains lower than the attention cost as soon as
\(L \gg D^{-1}\), which is the case in practice.

\paragraph{Exponential Causal Memory}

The memory is constructed by aggregating the \(N_b\) block summaries
through a normalized causal exponential average, followed by a linear
projection and a causal shift.

The overall complexity of these operations is

\[
\mathcal{O}(N_bD),
\]

that is, linear in the number of blocks.

\paragraph{Memory Injection}

Broadcasting the memory to the tokens of each block and the subsequent
gated fusion rely only on element-wise operations over representations of
dimension \(D\). Their computational cost is

\[
\mathcal{O}(TD).
\]

\paragraph{Overall Complexity}

The dominant computational cost of BCMT is the local attention, leading
to an overall complexity of

\[
\boxed{\mathcal{O}(TL)}.
\]

When the block size is fixed (\(L \ll T\)), this complexity becomes
linear with respect to the sequence length, whereas that of a Transformer
with global attention remains quadratic:

\[
\text{Dense Transformer}
\;:\;
\mathcal{O}(T^2),
\qquad
\text{BCMT}
\;:\;
\mathcal{O}(TL).
\]

BCMT therefore reduces both computational and memory costs by restricting
self-attention to local blocks, while long-range dependencies are
propagated through an exponential causal memory whose computational cost
is linear.

\section{Experiments}

\subsection{Experimental Setup}

\subsubsection{Evaluated Configurations}

All comparisons are conducted between a standard dense causal Transformer and several BCMT variants.

To isolate the impact of the proposed memory mechanism, all models share exactly the same architecture (number of layers, hidden dimension, and number of attention heads). The only difference between the Dense Transformer and BCMT lies in the strategy used to model dependencies.

We evaluate four BCMT variants corresponding to different block sizes:

\begin{center}
\begin{tabular}{lc}
\hline
Configuration & Block Size $L$ \\
\hline
BCMT-64  & 64  \\
BCMT-128 & 128 \\
BCMT-256 & 256 \\
BCMT-512 & 512 \\
\hline
\end{tabular}
\end{center}

\subsubsection{Model Architecture}

All models use the following configuration:

\begin{itemize}
\item hidden dimension: $d_{\text{model}} = 128$
\item number of layers: $6$
\item number of attention heads: $8$
\item batch size: $16$
\end{itemize}

All models are implemented in PyTorch and trained using the same hyperparameters to ensure a fair comparison.

\subsubsection{Datasets}

All experiments are conducted on the WikiText-103 corpus, a widely used benchmark for evaluating autoregressive language models.

Two training subsets are constructed from the training split in order to evaluate BCMT under two complementary experimental settings.

\paragraph{WikiText200k}

WikiText200k is obtained by randomly sampling 200,000 sequences from the training split\footnote{
\texttt{dataset["train"].shuffle(seed=999).select(range(200000))}
}.

After preprocessing and tokenization, this corpus contains approximately 12.60 million training tokens. It is used in Phases I, II, and IV.

\paragraph{WikiText600k}

WikiText600k is constructed using the same protocol by selecting 600,000 sequences\footnote{
\texttt{dataset["train"].shuffle(seed=123).select(range(600000))}
}.

After preprocessing and tokenization, this corpus contains approximately 38.04 million training tokens. It is used in Phase III to evaluate the behavior of BCMT under a more demanding training regime.

For all experiments, evaluation is performed on the entire WikiText-103 validation split, comprising 239,536 tokens.

\subsubsection{Training Protocol}

All models are trained in an autoregressive language modeling setting using a causal mask and the AdamW optimizer.

The learning rate is dynamically adjusted using a \textit{ReduceLROnPlateau} scheduler configured with:

\begin{itemize}
\item reduction factor: $0.5$;
\item patience: $2$ epochs;
\item improvement threshold: $10^{-3}$;
\item minimum learning rate: $10^{-5}$.
\end{itemize}

Training is terminated using an \textit{early stopping} criterion when the validation loss no longer improves, with:

\begin{itemize}
\item patience: $6$ epochs;
\item minimum required improvement (\textit{min\_delta}): $10^{-3}$.
\end{itemize}

Training sequences are generated with 50\% overlap. Each experiment is repeated using two independent random seeds (\texttt{seed=42} and \texttt{seed=123}). The reported results correspond to the average of the metrics obtained from these two runs, and the standard deviations are computed from the same initializations.

\subsubsection{Experimental Protocol}

The experimental evaluation is organized into four complementary phases.

\paragraph{Phase I: Sensitivity to Context Length}

Comparison between the Dense Transformer and BCMT-128 for sequence lengths

\[
\texttt{seq\_len}\in\{128,256,512,1024\},
\]

to evaluate the impact of context length on model performance and computational efficiency.

\paragraph{Phase II: Comparison of BCMT Variants}

Comparison between the Dense Transformer and the BCMT-64, BCMT-128, BCMT-256, and BCMT-512 variants to investigate the impact of block size on the trade-off between modeling quality and computational efficiency.

\paragraph{Phase III: Large-Scale Validation}

Comparison between the Dense Transformer and BCMT-256 on the WikiText600k corpus to verify that the modeling performance, computational efficiency gains, and robustness of the proposed memory mechanism are maintained as the amount of training data increases substantially.

\paragraph{Phase IV: Ablation Study}

Comparison between BCMT and a variant without inter-block memory (HOnly) to quantify the contribution of the proposed memory mechanism.


\subsection{Phase I: Sensitivity to Context Length}

\subsubsection{Objective}

The first experimental phase evaluates the behavior of BCMT as the context length increases.

Its objective is to assess BCMT's ability to preserve its modeling performance as the context length increases, while reducing the computational cost associated with global attention.

\subsubsection{Experimental Setup}

We compare the reference Dense Transformer with BCMT-128 for different context lengths:

\[
\texttt{seq\_len} \in \{128, 256, 512, 1024\}.
\]

All experiments are conducted on the WikiText200k corpus described previously.

We report the following metrics:

\begin{itemize}
    \item validation loss;
    \item validation perplexity;
    \item training throughput, measured in tokens per second;
    \item training time per epoch;
    \item peak GPU memory usage.
\end{itemize}

\subsubsection{Results}

\begin{table}[ht]
\centering
\caption{
Comparison of the scaling behavior with respect to context length between the Dense Transformer and BCMT-128.
Validation losses are reported as the mean $\pm$ standard deviation over two random seeds.
}
\label{tab:phase1_results}
\setlength{\tabcolsep}{4pt}
\begin{tabular}{lcccccc}
\hline
Model & Seq. Len & Val Loss $\downarrow$ & Val PPL $\downarrow$ & Tok/s $\uparrow$ & Time/Epoch $\downarrow$ & VRAM (GB) $\downarrow$ \\
\hline

Dense Transformer & 128  & $4.5396 \pm 0.0018$ & 93.65 & 105.8K & 238s & 1.47 \\
BCMT-128          & 128  & $4.5359 \pm 0.0038$ & 93.31 & 99.5K & 253s & 1.47 \\

\hline

Dense Transformer & 256  & $4.5261 \pm 0.0057$ & 92.40 & 149.9K & 168s & 2.89 \\
BCMT-128          & 256  & $4.5714 \pm 0.0035$ & 96.67 & 127.6K & 198s & 2.77 \\

\hline

Dense Transformer & 512  & $4.5330 \pm 0.0113$ & 93.04 & 154.1K & 164s & 6.16 \\
BCMT-128          & 512  & $4.5800 \pm 0.0046$ & 97.51 & 170.0K & 148s & 5.34 \\

\hline

Dense Transformer & 1024 & $4.5752 \pm 0.0063$ & 97.04 & 119.9K & 210s & 14.37 \\
BCMT-128          & 1024 & $4.6130 \pm 0.0069$ & 100.79 & 204.2K & 123s & 10.48 \\

\hline
\end{tabular}
\end{table}

Table~\ref{tab:phase1_results} reports the results of the context-length sensitivity study comparing the reference Dense Transformer with BCMT-128 for context lengths ranging from 128 to 1024 tokens. Validation losses are reported as the mean $\pm$ standard deviation over two independent random initializations.

The results show that BCMT maintains modeling performance close to that of the Dense Transformer across all evaluated context lengths while becoming progressively more computationally efficient as the context length increases.

For a context length of 128 tokens, the entire sequence fits within a single block, meaning that the inter-block memory mechanism is not activated. Under this setting, BCMT achieves a validation loss of $4.5359 \pm 0.0038$, slightly lower than that of the Dense Transformer ($4.5396 \pm 0.0018$). This result indicates that partitioning the sequence into blocks does not introduce any intrinsic degradation when the entire sequence is contained within a single block.

As the context length increases, the performance gap between the two architectures remains moderate. At a context length of 1024 tokens, BCMT achieves a validation loss of $4.6130 \pm 0.0069$, compared with $4.5752 \pm 0.0063$ for the Dense Transformer. Despite the absence of global attention across blocks, the exponential causal memory preserves a large portion of the information required for accurate modeling.

At the same time, the computational efficiency gains become increasingly pronounced as the context length grows. For a context length of 512 tokens, BCMT increases the training throughput from 154.1K to 170.0K tokens/s (+10.3\%) while reducing the peak GPU memory usage from 6.16 to 5.34~GB (-13.3\%). At 1024 tokens, these improvements become particularly significant: the training throughput reaches 204.2K tokens/s compared with 119.9K tokens/s for the Dense Transformer (+70.3\%), while the peak GPU memory usage decreases from 14.37 to 10.48~GB (-27.1\%).

These observations are consistent with the complexity analysis presented in Section~\ref{subsec:compcompu}. When the block size is fixed, the computational complexity of local attention grows linearly with the context length, whereas dense attention remains quadratic. Consequently, the computational advantages of BCMT become increasingly apparent as the context length increases.

Finally, the small standard deviations observed across all experiments indicate stable optimization. They show that the exponential causal memory mechanism does not introduce measurable optimization instabilities, regardless of the context length considered.

\subsection{Phase II: Comparison of BCMT Variants}

\subsubsection{Objective}

This second phase constitutes the primary evaluation of the paper.

Its objective is to investigate the impact of block size on the trade-off between modeling performance and computational efficiency in order to identify the most balanced configuration.

\subsubsection{Experimental Setup}

All experiments are conducted on the WikiText200k corpus with a fixed context length of

\[
\texttt{seq\_len}=1024.
\]

We compare the reference Dense Transformer with four BCMT variants:

\begin{itemize}
    \item BCMT-64;
    \item BCMT-128;
    \item BCMT-256;
    \item BCMT-512.
\end{itemize}

We report the following metrics:

\begin{itemize}
    \item validation loss;
    \item validation perplexity;
    \item training throughput (tokens/s);
    \item training time per epoch;
    \item peak GPU memory usage.
\end{itemize}

\subsubsection{Results}

\begin{table}[h]
\centering
\caption{
Comparison of the architectures evaluated in Phase II on WikiText200k with a context length of 1024 tokens.
Validation losses are reported as the mean $\pm$ standard deviation over two independent random seeds.
}
\label{tab:phase2_results}

\setlength{\tabcolsep}{4pt}

\begin{tabular}{lccccc}
\hline
Model &
Val Loss $\downarrow$ &
Val PPL $\downarrow$ &
Tok/s $\uparrow$ &
Time/Epoch $\downarrow$ &
VRAM (GB) $\downarrow$
\\
\hline

Dense Transformer
&
$4.5752 \pm 0.0045$
&
97.05
&
119.9K
&
210s
&
14.37
\\

BCMT-64
&
$4.6344 \pm 0.0037$
&
102.97
&
211.6K
&
119s
&
10.20
\\

BCMT-128
&
$4.6130 \pm 0.0049$
&
100.78
&
204.2K
&
123s
&
10.48
\\

BCMT-256
&
$\mathbf{4.5931 \pm 0.0045}$
&
$\mathbf{98.81}$
&
188.7K
&
134s
&
11.05
\\

BCMT-512
&
$4.5969 \pm 0.0030$
&
99.18
&
155.7K
&
162s
&
12.17
\\

\hline
\end{tabular}
\end{table}

Table~\ref{tab:phase2_results} compares the Dense Transformer with the different BCMT variants on WikiText200k using a fixed context length of 1024 tokens. Validation losses are reported as the mean $\pm$ standard deviation over two independent random initializations.

The results reveal a clear trade-off between modeling performance and computational efficiency, directly controlled by the block size. As the block size increases, predictive performance progressively approaches that of the Dense Transformer, while the gains in training throughput and memory efficiency gradually decrease.

Among the evaluated configurations, BCMT-256 provides the best trade-off. Its validation loss ($4.5931 \pm 0.0045$) remains very close to that of the Dense Transformer ($4.5752 \pm 0.0045$), with a gap of less than 0.02. At the same time, the training throughput increases from 119.9K to 188.7K tokens/s (+57.4\%), while the peak GPU memory usage decreases from 14.37 to 11.05~GB (-23.1\%).

At the opposite extreme, BCMT-64 maximizes computational efficiency. This configuration achieves the highest training throughput (211.6K tokens/s, corresponding to a +76.5\% improvement) and the lowest peak GPU memory usage (10.20~GB, a reduction of 29.0\%), at the cost of a more noticeable increase in validation loss ($4.6344 \pm 0.0037$). These results suggest that excessively small block sizes limit the amount of contextual information available locally.

The BCMT-128 and BCMT-512 variants illustrate intermediate points along this trade-off. BCMT-128 retains substantial computational gains (+70.4\% in training throughput) with a slightly larger degradation in modeling performance. Conversely, BCMT-512 achieves predictive performance very close to that of BCMT-256 while providing more modest computational gains (+29.8\% in training throughput). Increasing the block size therefore progressively improves modeling performance while reducing the computational benefits associated with local attention.

Overall, all BCMT variants provide substantial computational advantages over the Dense Transformer while maintaining comparable validation performance. These results show that the exponential causal memory effectively replaces global interactions across blocks with a compact contextual representation, without introducing a significant degradation in modeling performance.

Finally, the smooth evolution of performance as a function of block size confirms that this parameter provides an explicit means of controlling the trade-off between modeling performance and computational efficiency. Among the evaluated configurations, BCMT-256 offers the best balance between these two objectives.


\subsection{Phase III: Large-Scale Training Validation}

\subsubsection{Objective}

The third phase aims to verify that the benefits observed in the previous experiments are maintained as the amount of training data increases substantially.

\subsubsection{Experimental Setup}

The Dense Transformer and BCMT-256 are trained on the WikiText600k corpus with a fixed context length of

\[
\texttt{seq\_len}=1024.
\]

We report the following metrics:

\begin{itemize}
    \item validation loss;
    \item validation perplexity;
    \item training throughput (tokens/s);
    \item training time per epoch;
    \item peak GPU memory usage.
\end{itemize}

\subsubsection{Results}

\begin{table}[h]
\centering
\caption{
Results of Phase III on WikiText600k.
Validation losses are reported as the mean $\pm$ standard deviation over two independent random seeds.
}
\label{tab:phase3_results}

\begin{tabular}{lccccc}
\toprule
Model
&
Val Loss $\downarrow$
&
Val PPL $\downarrow$
&
Tok/s $\uparrow$
&
Time/Epoch $\downarrow$
&
VRAM (GB) $\downarrow$
\\
\midrule

Dense Transformer
&
$4.1115 \pm 0.0033$
&
61.04
&
120.2K
&
633s
&
14.37
\\

BCMT-256
&
$4.1470 \pm 0.0003$
&
63.24
&
189.7K
&
401s
&
11.05
\\

\bottomrule
\end{tabular}
\end{table}

The results reported in Table~\ref{tab:phase3_results} show that the trends observed in the first two phases are preserved when the amount of training data is tripled. Despite being trained on a corpus containing approximately 38 million tokens, BCMT retains a substantial computational advantage while maintaining modeling performance close to that of the Dense Transformer.

After training on WikiText600k, BCMT-256 achieves a validation loss of $4.1470 \pm 0.0003$, compared with $4.1115 \pm 0.0033$ for the Dense Transformer. Although the latter retains a slight advantage in modeling performance, the gap remains small relative to the computational gains achieved by BCMT.

The training throughput increases from 120.2K to 189.7K tokens/s, corresponding to a gain of 57.8\%, while the training time per epoch decreases from 633 to 401 seconds (-36.7\%). At the same time, peak GPU memory usage is reduced from 14.37 to 11.05~GB (-23.1\%).

The observed standard deviations remain very small for both architectures, and particularly for BCMT-256 ($\pm0.0003$), indicating excellent training stability despite the substantial increase in training data.

This third study confirms that the benefits of BCMT are not limited to a moderate-scale training regime. The improvements in training throughput and memory efficiency are preserved as the amount of training data increases substantially, demonstrating that the exponential causal memory mechanism remains both effective and stable under a more demanding training regime.


\subsection{Phase IV: Ablation Study}

\subsubsection{Objective}

The final phase aims to quantify the contribution of the memory mechanism proposed in BCMT.

More specifically, we compare the complete architecture with a simplified variant, denoted \textit{HOnly}, in which both the adaptive block summaries and the exponential causal memory are entirely removed.

The objective is to determine whether the improvements observed in the previous phases genuinely arise from the proposed inter-block memory mechanism or merely from partitioning the sequence into local blocks.

\subsubsection{Experimental Setup}

All experiments are conducted on the WikiText200k corpus with a fixed context length of

\[
\texttt{seq\_len}=1024.
\]

For each block size

\[
L \in \{64, 128, 256, 512\},
\]

we compare:

\begin{itemize}
\item \textbf{BCMT}, which uses adaptive block summaries and exponential causal memory;
\item \textbf{HOnly}, in which each block is processed independently without adaptive summaries or inter-block memory.
\end{itemize}

In the HOnly variant, the local representations are simply concatenated after local attention:

\[
H=
\operatorname{Concat}
\left(
H^{(1)},\ldots,H^{(N_b)}
\right),
\]

without any information propagation across blocks.

\subsubsection{Results}

\begin{table}[ht]
\centering
\caption{
Comparison between BCMT and its HOnly variant.
The gain corresponds to the difference in validation loss between HOnly and BCMT.
A positive value indicates an advantage in favor of BCMT.
}
\label{tab:ablation_honly}
\setlength{\tabcolsep}{5pt}
\begin{tabular}{lccc}
\hline
Block Size & BCMT & HOnly & Gain \\
\hline

64  & 4.6344 & 4.7026 & +0.0682 \\
128 & 4.6130 & 4.6516 & +0.0386 \\
256 & 4.5931 & 4.6052 & +0.0121 \\
512 & 4.5969 & 4.6024 & +0.0055 \\

\hline
\end{tabular}
\end{table}

Table~\ref{tab:ablation_honly} compares BCMT with its HOnly variant, in which the inter-block memory mechanism is removed. The gain corresponds to the difference in validation loss between HOnly and BCMT; a positive value indicates an advantage in favor of BCMT.

Removing the inter-block memory consistently increases the validation loss, regardless of the block size. This observation demonstrates that the improvements achieved by BCMT do not arise solely from partitioning the sequence into local blocks, but from the contextual propagation mechanism provided by the exponential causal memory.

However, the impact of this memory strongly depends on the block size. For blocks of 64 tokens, removing the memory increases the validation loss by 0.0682. The improvement remains substantial for a block size of 128 tokens (+0.0386), but decreases markedly as the block size increases (+0.0121 for 256-token blocks and only +0.0055 for 512-token blocks).

This behavior is consistent with the underlying design of BCMT. When the blocks are small, local attention covers only a limited portion of the context, making the inter-block memory essential for propagating information from preceding blocks. As the block size increases, an increasing fraction of the dependencies is captured directly by local attention, naturally reducing the contribution of the memory mechanism.

These results therefore highlight the complementarity of the two components of the architecture. Local attention models short-range dependencies, while the exponential causal memory effectively compensates for the loss of context induced by block-wise processing when this limitation is most severe.

Finally, this ablation study validates the central hypothesis underlying BCMT. The observed performance gains do not result solely from local block processing, but from the combination of local attention and inter-block memory. The proposed memory mechanism is therefore the key component that preserves long-range dependencies while maintaining the computational efficiency benefits of localized attention.


\section{Conclusion}

We have presented BCMT (\textit{Blockwise Causal Memory Transformer}), a long-context language modeling architecture based on an explicit separation between local interactions and global context propagation. While interactions between tokens are modeled through dense causal self-attention applied independently within local blocks, long-range dependencies are propagated by means of an exponential causal memory constructed from adaptive block summaries.

This design enables the propagation of global contextual information without relying on explicit attention between all tokens in the sequence, while preserving a simple, fully parallelizable architecture that remains compatible with standard dense self-attention implementations.

Experiments on WikiText show that BCMT achieves validation performance comparable to that of a Dense Transformer while significantly improving training throughput and reducing GPU memory consumption. The ablation study confirms that these improvements arise from the proposed memory mechanism rather than from local block processing alone, highlighting the essential role of context propagation across blocks.

These results demonstrate that a compact memory constructed from adaptive block summaries constitutes a viable alternative to dense global attention for long-context language modeling. More broadly, they suggest that explicitly separating the modeling of local interactions from the propagation of global context is a promising architectural principle for developing more efficient sequence models.

Several directions for future research can be envisioned. In particular, it would be interesting to investigate more expressive summarization mechanisms, hierarchical or multi-scale memory structures, as well as the application of BCMT to larger models and tasks requiring stronger reasoning capabilities or long-range information retrieval.

We hope that this work will contribute to the development of a new family of architectures in which local modeling and global context propagation are handled by distinct yet complementary mechanisms, providing a simple, efficient, and fully parallelizable alternative to architectures based on dense global attention.

\section{Limitations}

BCMT propagates long-range dependencies through an exponential causal memory constructed from adaptive block summaries rather than through explicit interactions between all tokens in the sequence. This design choice significantly reduces the computational cost, but relies on a deliberately compact representation of the historical context.

By construction, this memory is not intended to preserve a detailed representation of every observed token. The exponential causal aggregation naturally emphasizes the most recent information while maintaining a progressively decreasing contribution from earlier blocks. Although this strategy provides an effective global contextual signal, it does not allow for the exact retrieval of information from the distant past. Tasks requiring precise localization or faithful reconstruction of distant events may therefore continue to benefit from explicit global attention mechanisms or specialized memory systems.

Furthermore, the experimental evaluation presented in this work is intentionally focused. The experiments are primarily conducted on WikiText language modeling tasks and compare BCMT with a standard Dense Transformer in order to isolate the contribution of the proposed memory mechanism. While this methodology enables a precise assessment of BCMT itself, it does not encompass the full range of recent long-context architectures. A systematic comparison with models such as Transformer-XL, Recurrent Memory Transformer (RMT), RetNet, RWKV, and Mamba would constitute an important next step for positioning BCMT more precisely within the current landscape of memory-augmented architectures.

Finally, the current version of BCMT relies on a single memory level and a fixed exponential decay factor. Hierarchical, multi-scale, or adaptive memory mechanisms could provide a more effective representation of dependencies spanning different temporal ranges while preserving linear computational complexity. Exploring such extensions constitutes a natural direction for future work.


\bibliography{references}

\end{document}